\documentclass{article}

\usepackage{arxiv}

\usepackage[utf8]{inputenc} 
\usepackage[T1]{fontenc}    
\usepackage{hyperref}       
\usepackage{url}            
\usepackage{booktabs}       
\usepackage{amsfonts}       
\usepackage{nicefrac}       
\usepackage{microtype}      
\usepackage{subcaption}
\usepackage{lipsum}
\usepackage{graphicx}
\usepackage[numbers,sort&compress]{natbib}
\graphicspath{ {./images/} }

\title{LiLa-WAM: Lightweight Latent Reasoning World-Action Model for Robotic Manipulation}

\author{%
\begin{tabular}{c}
Fan Yang\textsuperscript{1} \quad
Yuting Su\textsuperscript{1} \quad
Xiaobo Wang\textsuperscript{2,5} \quad
Yuncheng You\textsuperscript{1} \quad
Fugui Fan\textsuperscript{1} \\
Yuting Wu\textsuperscript{4} \quad
Minghui Wu\textsuperscript{3} \quad
Chenxu Zhao\textsuperscript{3} \quad
Jiahong Ning\textsuperscript{5} \quad
Peiguang Jing\textsuperscript{1} \\[0.45em]
\textsuperscript{1}Tianjin University \quad
\textsuperscript{2}Shenzhen University of Advanced Technology \quad
\textsuperscript{3}Mininglamp Technology \\
\textsuperscript{4}Ministry of Natural Resources Information Center \quad
\textsuperscript{5}Sangfor Technologies Inc. \\
\end{tabular}
}

\begin{document}
\pagestyle{plain}

\maketitle
\begin{abstract}
World-action modeling has emerged as a promising paradigm for robotic control, as it empowers models to go beyond reacting to observations and anticipate how a scene will evolve. 
However, existing WAMs often incur substantial computational overhead. 
Pixel-space methods often allocate substantial capacity to visual details that may not be directly relevant to control, while some latent-space methods require multi-stage training to construct the reasoning space. The resulting training cost can make such methods difficult to train under modest computational budgets.
In this work, we propose LiLa-WAM, a lightweight world-action model that reasons about the future in a compact latent space and can be trained end-to-end on a single 24GB GPU. 
Its core design is a compact latent reasoning space jointly shaped by future-state prediction and action generation, which keeps the model lightweight while remaining well aligned with control. 
For task specification, we further propose the Visual Transition Token(VTT), a language-free task representation that encodes each task as a direction in visual feature space.
Experiments on RoboTwin~2.0, LIBERO, and real-robot tasks demonstrate LiLa-WAM's effectiveness, achieving 90.48\% success across 50 RoboTwin tasks with single-GPU training.
Code is available at \url{https://github.com/teee000/LiLa-WAM}.

\end{abstract}


\section{Introduction}

World-Action Models (WAMs)~\cite{vpp,unipi,seer,wall_wm,aha_wam} have emerged as a promising paradigm for robotic manipulation.
By unifying predictive state
modeling with action generation, WAMs enable the policy to anticipate how a scene will evolve under intervention, rather than merely react to current observations~\cite{survey_wam2026}.
Such foresight can benefit robust control in manipulation tasks.

Existing WAMs can be broadly grouped by the representation space in which the future is predicted. 
\textbf{Pixel-space methods} explicitly synthesize future observations as images or videos~\cite{imagewam,pi07,cosmos_policy,lingbot_va,fast_wam}. 
However, generating dense future frames, whether a single frame or a video sequence, tends to allocate substantial capacity to control-irrelevant details such as texture, lighting, and background, making both training and inference
computationally expensive and memory-intensive. \textbf{Latent-space methods}
instead predict future states in a latent representation
space~\cite{lapa,flare,frappe,wog,beingh07}, which avoids such appearance redundancy.
Nevertheless, a number of these methods construct their latent reasoning space
through a multi-stage pipeline, where key components are often trained
separately and subsequently integrated with the policy, further increasing the
overall training burden.

\begin{figure}[t]
\centering
\begin{subfigure}[t]{0.48\textwidth}
    \centering
    \includegraphics[width=\linewidth, trim=5 5 5 5, clip]{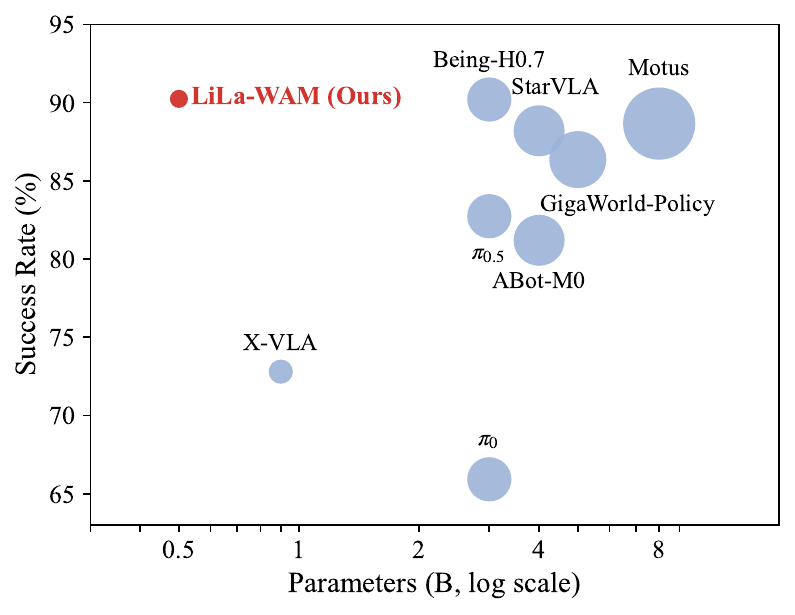}
    \caption{Comparison on RoboTwin~2.0 (50 tasks).}
    \label{fig:compare_robotwin}
\end{subfigure}
\hfill
\begin{subfigure}[t]{0.48\textwidth}
    \centering
    \includegraphics[width=0.985\linewidth, trim=5 5 5 5, clip]{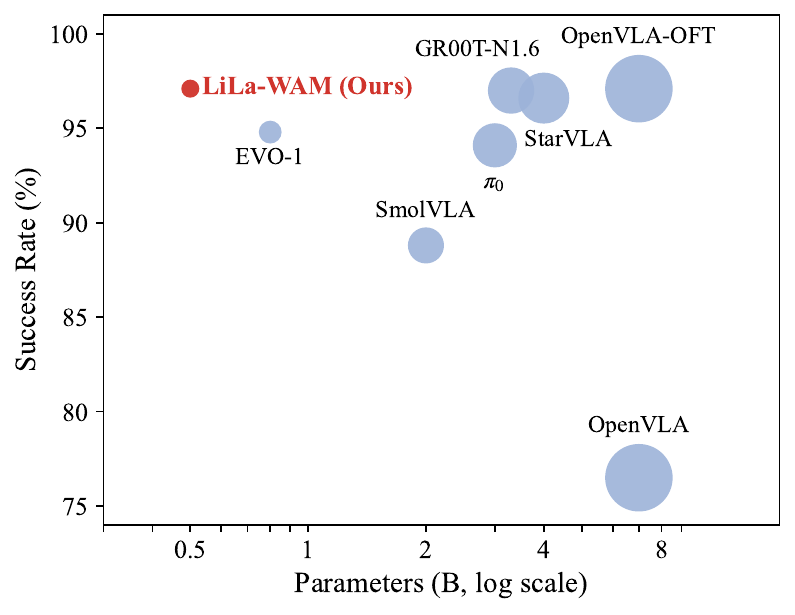}
    \caption{Comparison on LIBERO.}
    \label{fig:compare_libero}
\end{subfigure}
\caption{Average success rate versus model size on RoboTwin~2.0 and LIBERO, for methods with reported model sizes. LiLa-WAM achieves competitive average performance while using substantially fewer parameters than most compared methods, maintaining a lightweight design.}
\label{fig:compare_datasets}
\end{figure}

Despite this rapid progress, current WAMs remain expensive to train, limiting their accessibility under modest computational budgets.
The primary source of the
computational burden lies in the backbones. Current WAMs, regardless of their prediction space, are generally built upon large-scale pretrained models, whose capacity is largely devoted to functionalities beyond robotic control.
For vision-language backbones, a considerable portion of parameters serves
language modeling rather than physical interaction. For video generation
backbones, substantial capacity is spent on synthesizing visual details that are not tightly coupled with action control. 
These observations suggest that compact, detail-rich visual backbones may
offer a viable foundation for robot control, motivating a lightweight
world-action model that relies solely on a visual backbone and receives task
information from visual features rather than language.

In this work, we propose LiLa-WAM, a lightweight latent-reasoning
world-action model for robotic manipulation.
At its core, the Foresight-Aware Action Expert unifies future-state prediction and action generation within a single stream, allowing the two objectives to jointly shape a compact latent space.
Beyond language instructions, recent works have explored specifying tasks through visual cues, such as interleaved image-text instructions or goal images~\cite{pi07}.
Along this line, we take a further step toward simplicity by introducing the Visual Transition Token (VTT), a language-free task representation computed from the visual embeddings of task demonstrations. The VTT encodes each task as a transition direction in feature space to capture what the task changes about the scene. Compared with goal-image conditioning, it requires neither text nor a goal image at test time, providing a lightweight alternative for task specification.
We evaluate LiLa-WAM on RoboTwin~2.0 (50 tasks with a single model), LIBERO, and real-robot tasks, and analyze its key design choices through extensive ablation studies and visualizations. Our contributions are summarized as follows:
\begin{itemize}
\item We present {LiLa-WAM}, a lightweight world-action model that predicts future states and generates actions in a compact latent space, and can be trained end-to-end on a single 24\,GB GPU.
\item We propose the {Visual Transition Token (VTT)}, a language-free task representation that encodes each task as a transition direction in visual feature space, without requiring text or goal images at test time.
\item We empirically evaluate LiLa-WAM across simulation and real-world settings, where it remains competitive with larger policies.
\end{itemize}

\begin{figure*}[t]
\centering
\includegraphics[width=\textwidth, trim=20 264 210 5,clip]{ 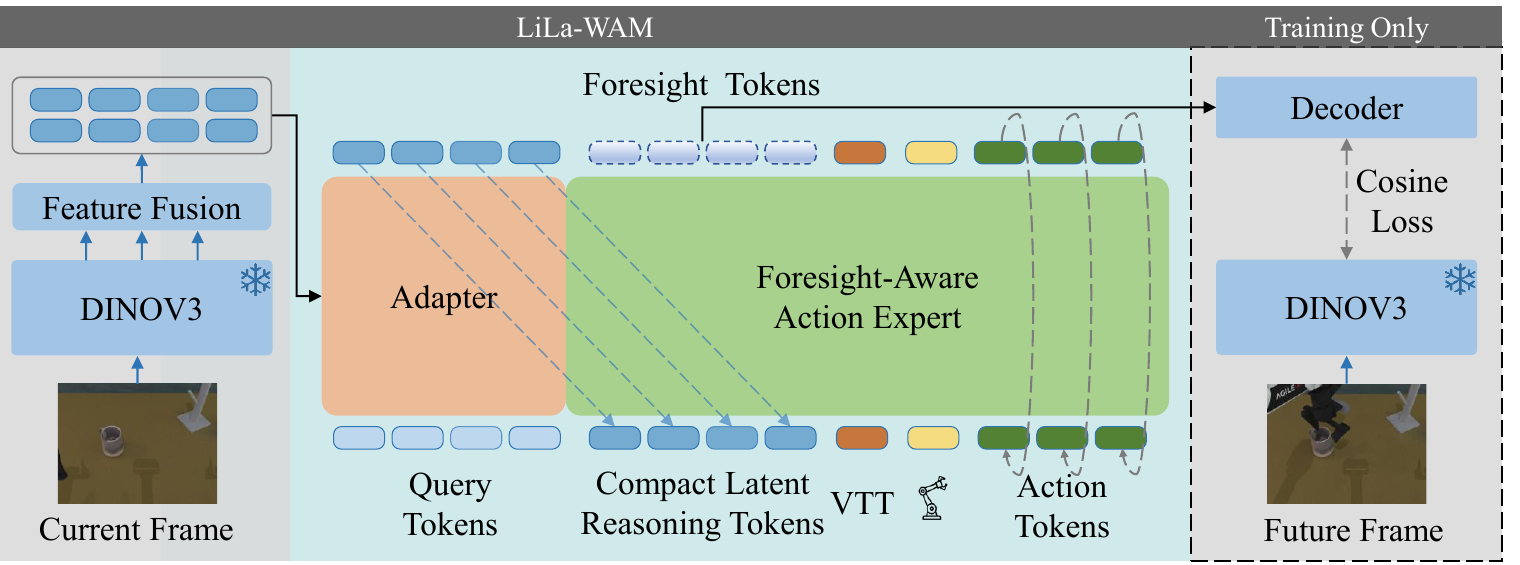} 
\caption{Overview of LiLa-WAM. The Foresight-Aware Action Expert unifies
reasoning tokens, the VTT, proprioceptive tokens, and noisy action tokens
in a single stream, jointly producing the action velocity and foresight
tokens, which are supervised in the feature space during training
and discarded at inference.}
\label{fig:Framework}
\end{figure*}

\section{Related Work}
\label{sec:related}

\subsection{Vision-Language-Action Models}
Vision-language-action (VLA) models transfer web-scale vision-language
pretraining into robot control~\cite{openvla,octo,pi0}. Diffusion- or
flow-matching-based policies~\cite{dp,rdt,cogact} model continuous action
chunks and have become the dominant architecture. A recent trend pursues
lightweight VLAs that reduce backbone size and training
cost~\cite{tinyvla,smolvla,evo1}. Despite their differences, these models
generally map the current observation directly to actions and remain largely
reactive: how the scene will evolve under the robot's actions is typically
left unmodeled, and whether such foresight can emerge from demonstrations
alone remains unclear. LiLa-WAM shares the lightweight pursuit but differs
fundamentally in formulation. It couples action generation with explicit
future-state prediction in a compact latent space, so that foresight becomes
a training signal rather than an emergent by-product.

\subsection{World-Action Models}
World-action models (WAMs) extend VLAs with the ability to predict future
states alongside action generation. A first line operates in pixel space.
Imagine-then-act methods~\cite{unipi,mimic_video,vidar,vpp,genie} first
generate future frames or videos and then recover actions through inverse
dynamics or a future-conditioned policy, which decouples prediction from
control at the cost of compounding errors and deployment latency. Joint
video-action methods~\cite{worldvla,uva,motus,lingbot_va,gigaworld_policy,aha_wam}
instead model future frames and action chunks within a shared generative
architecture, typically initialized from video-generation backbones such as
WAN~\cite{wan}. While effective, the appearance details these models learn
to synthesize are often not strongly related to control, and training or
adapting such generative backbones demands computational budgets far beyond
a single GPU.

A second line performs future prediction in latent space~\cite{flare,frappe,wog,lawam}. WoG~\cite{wog}
learns an action-oriented condition space by first grounding future
observations into the action pipeline and then distilling them into the
VLM. LaWAM~\cite{lawam} infers latent actions from visual transitions and trains
a latent world model that decodes them into future features, which is
integrated into a VLA policy in a second stage to provide latent visual
subgoals. Being-H0.7~\cite{beingh07} shapes its
latent space with a posterior branch that accesses future observations
during training, relying on additional alignment losses to prevent feature
collapse. These methods avoid pixel synthesis, but introduce their own
complexities: WoG and LaWAM construct the reasoning space through two-stage
pipelines that couple separately trained modules to the policy only
afterwards, while Being-H0.7 requires auxiliary objectives beyond direct
future supervision to keep the latent space well-behaved.
LiLa-WAM also predicts the future in latent space, but integrates
future-state prediction and action generation into a single stream trained
end-to-end.

\subsection{Visual Task Specification}
Beyond the dominant channel of language instructions, recent works have
explored visual alternatives for task
specification~\cite{act2goal,goal_vla,gevrm}. Goal-image and subgoal-image
conditioning~\cite{pi07,rt_sketch} specify the task through a target visual
state, yet require a concrete goal frame at test time. VIP~\cite{vip}
pretrains policies to interpret an instruction image depicting the desired
interaction, and Interleave-VLA~\cite{interleave_vla} shows that interleaved
image-text instructions substantially improve out-of-distribution
generalization over text-only conditioning, at the price of heavier
multimodal inputs and instruction construction. Our Visual Transition Token
takes a further step toward simplicity: it encodes the task as a transition
direction in visual feature space computed offline from demonstrations,
requiring neither text, nor paired instruction construction, nor a goal
frame at deployment, while empirically concentrating on task-relevant
objects.

\section{Method}
\label{sec:method}

\subsection{Problem Formulation}
\label{sec:formulation}

We consider robot manipulation from visual observations. At time step $t$,
the robot receives an image observation $o_t$ and a proprioceptive state
$q_t$, and predicts an action chunk $\mathbf{a}_{t:t+H} = (a_t, a_{t+1},
\dots, a_{t+H-1})$ over horizon $H$. A standard vision-language-action policy
models the conditional distribution $p(\mathbf{a}_{t:t+H} \mid o_t, q_t, l)$,
mapping the current observation and a language instruction $l$ directly to
actions. Such a formulation focuses mainly on reacting to the current observation, with limited explicit modeling of how the scene will evolve once the robot acts.

World-action models address this limitation by modeling future states and
actions jointly rather than actions alone. Following this principle, LiLa-WAM
learns
\begin{equation}
\label{eq:joint}
p_\theta\big(\mathbf{a}_{t:t+H},\, \hat{\mathbf{z}}_{t+\Delta} \mid o_t, q_t, c_\tau\big),
\end{equation}
where $\hat{\mathbf{z}}_{t+\Delta}$ is the predicted compact latent of the
future state after a look-ahead interval $\Delta$, $\theta$ denotes all
trainable parameters, and $c_\tau$ is a visual task cue that replaces the
language instruction. Here the future is not represented
by raw observations or pixel reconstructions, but by a compact latent
produced within the action expert itself, which keeps Eq.~\ref{eq:joint}
free from any heavy generative backbone.

\subsection{Foresight-Aware Action Expert}
\label{sec:expert}
\noindent\textbf{Motivation.}
The design of the action expert is guided by one goal: achieving lightweight
yet foresight-aware world-action modeling. Existing WAMs fall short of this
goal in different ways. Pixel-space methods synthesize future observations
explicitly, spending substantial capacity on appearance details and relying
on heavy generative backbones. Latent-space methods avoid pixel synthesis,
but often construct their reasoning space through a multi-stage pipeline,
where the predictive components are trained separately and coupled to the
policy only afterwards.

A deeper source of overhead lies in a mismatch between pretraining
objectives and the demands of robot control. VLM-based VLA models inherit
backbones pretrained with next-token prediction or image-text contrastive
alignment on web-scale data~\cite{qwen3vl,paligemma}, objectives that favor
high-level semantics over the fine-grained spatial and geometric cues that
precise manipulation requires; video-generation-based WAMs likewise devote
most of their parameters to modeling pixel-level dynamics rather than
control-relevant state. In both cases, a substantial portion of the model's capacity and computation may not be allocated to the cues most relevant to fine-grained control. We therefore build on DINOv3~\cite{dinov3}, a self-supervised visual encoder
whose dense pretraining yields patch features rich in fine-grained visual detail, and concentrate all trainable capacity in a
single lightweight stream in which future-state prediction and action
generation share and jointly shape one compact latent space.

\noindent\textbf{Architecture.}
We build LiLa-WAM on the frozen DINOv3 encoder $\mathcal{V}$ and compress
its dense patch features into a small set of tokens with a query-based
adapter $\mathcal{A}$,
\begin{equation}
\label{eq:adapter}
\mathbf{Z}_v
= \mathcal{A}\big(\mathbf{Q},\, \tilde{\mathcal{V}}(o_t)\big)
\in \mathbb{R}^{N_q \times D},
\end{equation}
where $\mathbf{Q} \in \mathbb{R}^{N_q \times D}$ is a set of learnable queries
that cross-attends to the patch features, $\mathbf{Z}_v$ is the compressed
visual latent, $\tilde{\mathcal{V}}(o_t)\in\mathbb{R}^{N_p\times D}$ denotes the fused multi-level features, $N_q$ is the number of queries, and $D$ is the hidden width of
the action expert. Prior studies on depth estimation and fine-grained
recognition have shown that features from different backbone layers encode
complementary information at different granularities~\cite{depth_anything,dinov3,fpn,chen2022vision}.
Since precise action prediction likewise depends on such fine-grained cues,
we exploit multi-level visual context before compression. Concretely, we
concatenate the patch features of several layers of $\mathcal{V}$ along the
channel dimension and project the result back to the expert width $D$ with a
linear layer, so that the adapter receives a fusion of low-level
spatial detail and high-level semantics at every patch. The adapter $\mathcal{A}$ stacks interleaved self-attention and
cross-attention layers, and compresses the fused features into $N_q$ tokens, yielding a compact latent space for foresight-aware
reasoning. Since $N_q$ is fixed regardless of the input resolution, the latent size is
decoupled from the patch count, bounding the cost of all downstream
attention.

The action expert $v_\theta$ is a stack of $L$ Diffusion-Transformer (DiT)
blocks. At each flow-matching step, its input is a single token sequence
\begin{equation}
\label{eq:seq}
\big[\, \mathbf{Z}_a;\; \mathbf{Z}_q;\; c_\tau;\; \mathbf{Z}_v \,\big],
\end{equation}
where $\mathbf{Z}_a \in \mathbb{R}^{H \times D}$ embeds the noised action
chunk $\mathbf{x}_s$, $\mathbf{Z}_q$ embeds the proprioceptive state $q_t$,
$c_\tau$ is the VTT-derived task token, and $\mathbf{Z}_v$ is the
compressed visual latent from Eq.~\ref{eq:adapter}. After the final block, a linear head reads out the predicted velocity
$\hat{\mathbf{v}}_s\in\mathbb{R}^{H\times d_a}$ at the action-token
positions, while the observation-side outputs form the predicted future
latent $\hat{\mathbf{z}}_{t+\Delta}\in\mathbb{R}^{N_q\times D}$.

\begin{figure}[t]
\centering
\begin{subfigure}[t]{0.48\textwidth}
    \centering
    \includegraphics[width=\linewidth, trim=5 295 560 5, clip]{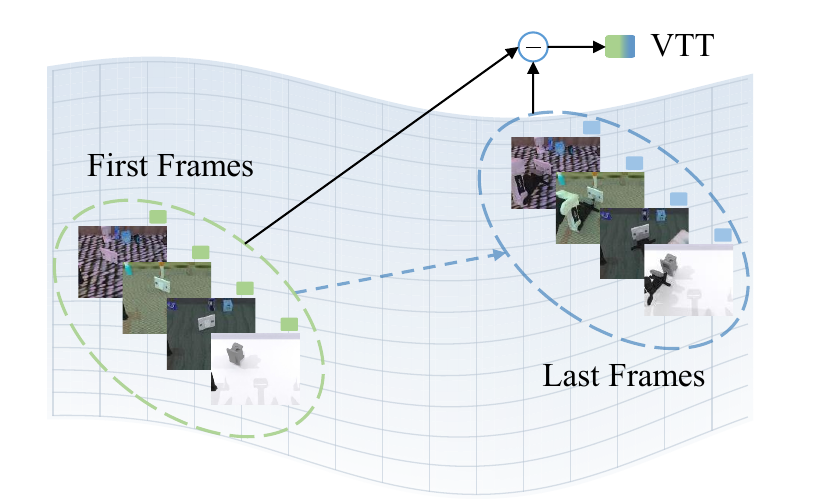}
    \caption{Illustration of the VTT, computed as the mean difference between last- and first-frame embeddings.}
    \label{fig:VisionTransToken}
\end{subfigure}
\hfill
\begin{subfigure}[t]{0.48\textwidth}
    \centering
    \includegraphics[width=\linewidth, trim=5 300 590 5, clip]{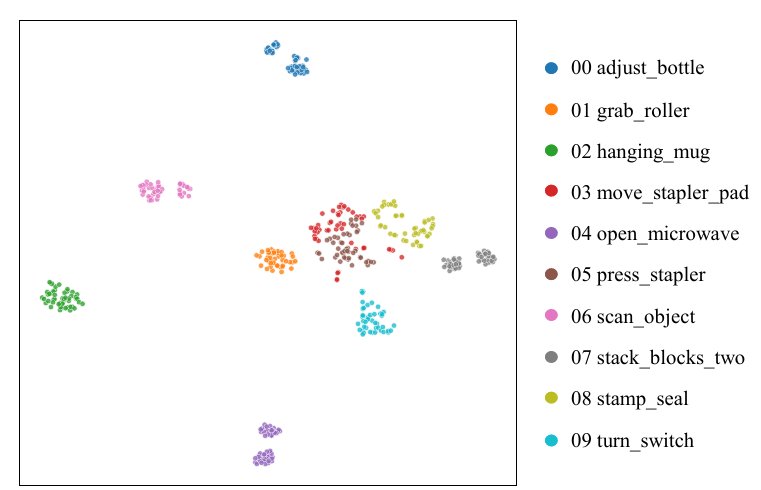}
    \caption{t-SNE of per-episode visual transition embeddings on 10 RoboTwin~2.0 tasks (clean and randomized).}
    \label{fig:TsneRobotwin}
\end{subfigure}
\caption{The Visual Transition Token and its embedding structure on RoboTwin~2.0.}
\label{fig:vtt_and_tsne}
\end{figure}

\noindent\textbf{Future-state prediction.}
During training, $\hat{\mathbf{z}}_{t+\Delta}$ is decoded back to the feature
space of the frozen encoder through a lightweight query-based (Q-Former-style)
decoder $\mathcal{D}$, and is supervised there against the patch features
that $\mathcal{V}$ extracts from the ground-truth future observation
$o_{t+\Delta}$. In this way, future prediction
serves as an auxiliary objective that shapes the shared latent, while no
pixel decoder or heavy generative module is ever required. Because
$\hat{\mathbf{z}}_{t+\Delta}$ and the action velocity are read out from the
same token sequence, gradients from the two objectives flow into one shared
representation, which encourages the shared latent to remain compact and control-relevant.
At inference, the expert inherently produces the future-aware latent together
with actions in a single forward pass, and the decoder is discarded,
introducing no extra test-time cost.

\subsection{Visual Transition Token}
\label{sec:vtt}
\noindent\textbf{Motivation.}
Manipulation policies typically specify the task with a language
instruction, incurring text-processing overhead. We instead seek a task
representation that is language-free, cheap to compute, and available
without any goal frame during deployment.

For a task $\tau$, let $\mathcal{E}_\tau$ be its set of demonstration
episodes. For each episode $e$, we take the global image embeddings of its
initial and final frames from the same frozen backbone, denoted
$\mathbf{g}^{e}_{0}$ and $\mathbf{g}^{e}_{T}$, and define the Visual
Transition Token as their mean difference over the task,
\begin{equation}
\label{eq:vtt}
\mathbf{u}_\tau
= \frac{1}{|\mathcal{E}_\tau|}
\sum_{e \in \mathcal{E}_\tau}
\big( \mathbf{g}^{e}_{T} - \mathbf{g}^{e}_{0} \big),
\end{equation}
where $|\mathcal{E}_\tau|$ is the number of episodes of task $\tau$. The VTT
encodes the task as a single transition direction in feature space, i.e.,
\emph{what the task changes about the scene}. The VTT is projected to the expert width with a lightweight MLP and
injected into Eq.~\ref{eq:seq} as the task token $c_\tau$, a fixed
per-task vector that requires neither text nor a goal frame at test time.

\subsection{Training Objective}
\label{sec:objective}
LiLa-WAM is trained end-to-end with two complementary objectives: a flow-matching
loss $\mathcal{L}_{\mathrm{fm}}$ for action-chunk prediction and a foresight
loss $\mathcal{L}_{\mathrm{ff}}$ for future-state prediction,
\begin{equation}
\label{eq:total}
\mathcal{L}
= \mathcal{L}_{\mathrm{fm}}
+ \lambda_{\mathrm{ff}}\,\mathcal{L}_{\mathrm{ff}},
\end{equation}
where $\lambda_{\mathrm{ff}}$ balances the two terms.

For action prediction, we adopt conditional flow matching. Let
$\mathbf{x}_0 \sim \mathcal{N}(\mathbf{0}, \mathbf{I})$ be a noise sample,
$\mathbf{x}_1$ the ground-truth action chunk, and $s \in [0,1]$ the flow
time. We form the linear interpolation $\mathbf{x}_s = (1-s)\,\mathbf{x}_0 +
s\,\mathbf{x}_1$ with the constant target velocity $\mathbf{v}^\star =
\mathbf{x}_1 - \mathbf{x}_0$, and train the expert $v_\theta$ to predict this
velocity,
\begin{equation}
\label{eq:fm}
\mathcal{L}_{\mathrm{fm}}
= \mathbb{E}_{\mathbf{x}_0,\,s}\;
\big\| \hat{\mathbf{v}}_s - \mathbf{v}^\star \big\|_2^2,
\end{equation}
where the conditioning set $\mathcal{C} = \{o_t, q_t, c_\tau\}$ enters through
the token sequence in Eq.~\ref{eq:seq}, and the velocity is read out at the
output positions corresponding to the action tokens $\mathbf{Z}_a$.

Future-state prediction is supervised in the DINOv3 feature space.
Specifically, a lightweight Q-Former-style decoder $\mathcal{D}$ maps the
predicted future latent $\hat{\mathbf{z}}_{t+\Delta}$ to the patch-feature
space of the frozen encoder. The foresight loss
$\mathcal{L}_{\mathrm{ff}}$ is computed as a per-token cosine loss between
the decoded features and the patch features extracted by $\mathcal{V}$ from
the ground-truth future observation $o_{t+\Delta}$. The decoder is used only
during training and is discarded at inference.

\section{Experiments}

\subsection{Implementation Details}
\label{sec:impl}

LiLa-WAM builds on a frozen DINOv3-ViT-L/16 encoder~\cite{dinov3}, using features from its 17th and 21st blocks. The visual features are processed by a 64-query adapter with 4 layers, followed by the Foresight-Aware Action Expert, a 12-layer DiT with 8 attention heads and a feature dimension of 768, and a 4-layer decoder. The policy takes a 16-dimensional proprioceptive state as input and predicts a 14-dimensional action sequence with a chunk length of 32, of which the first 16 steps are executed. The camera resolution is $320 \times 240$. The future horizon $\Delta$ is set to the action chunk length, i.e., $\Delta = 32$. The model contains 0.5B parameters, of which 0.2B are trainable and 0.3B are frozen.

We train with AdamW~\cite{adamw} ($\beta_1 = 0.9$, $\beta_2 = 0.99$, weight decay 0.01), a batch size of 128, gradient clipping with a maximum norm of 1.0, and $\lambda_{\mathrm{ff}} = 0.5$. During flow-matching training, the timestep is sampled from a logit-normal distribution. We adopt a stage-wise cosine learning-rate schedule, using a peak learning rate of $2 \times 10^{-4}$ in the early stage and $5 \times 10^{-5}$ in the later stage. All training fits within a single 24\,GB GPU budget: joint training over the 50 RoboTwin~2.0 tasks takes about 110 GPU hours on an NVIDIA RTX 5090.

At deployment, the VTT corresponding to the target task is selected, and action chunks are generated by integrating the learned velocity field with the Euler method using 10 ODE steps. Inference takes 85\,ms on an NVIDIA RTX 4090.

\subsection{Simulation Environments}
\label{sec:simenv}

\paragraph{RoboTwin~2.0.}
RoboTwin~2.0~\cite{robotwin2} serves as our primary benchmark, covering 50
manipulation tasks. For each task, we collect 50 clean demonstrations and
500 randomized demonstrations with domain randomization, and train a single
model jointly on the data of all tasks. The policy receives a single
third-person RGB view resized to $320 \times 240$, and predicts action
chunks of size 32. Each task is evaluated over 50 rollouts. For ablation
studies, we additionally construct a subset of 10 representative tasks
spanning different types and difficulty levels:
\textit{adjust\_bottle}, \textit{grab\_roller}, \textit{hanging\_mug},
\textit{move\_stapler\_pad}, \textit{open\_microwave},
\textit{press\_stapler}, \textit{scan\_object}, \textit{stack\_blocks\_two},
\textit{stamp\_seal}, and \textit{turn\_switch}.

\paragraph{LIBERO.}
We further evaluate on LIBERO~\cite{libero} to compare with lightweight VLA
baselines. The policy takes both a third-person view and a wrist-mounted
view resized to $256 \times 256$, and predicts action chunks of size 12.
A separate policy is trained for each of the four suites, each containing
10 tasks (40 tasks in total), and each task is evaluated over 50 rollouts.

\begin{table}[t]
\centering
\begin{minipage}[t]{0.43\textwidth}
    \centering
    \small
    \setlength{\tabcolsep}{1mm}
    \begin{tabular}{llcc}
      \toprule
      Method & Params & Clean & Random \\
      \midrule
      ABot-M0 & 4B & 81.20 & 80.40 \\
      StarVLA & 4B & 88.20 & 88.30 \\
      $\pi_0$ & 3B & 65.92 & 58.40 \\
      $\pi_{0.5}$ & 3B & 82.74 & 76.76 \\
      X-VLA & 0.9B & 72.88 & 72.84 \\
      \midrule
      Motus & 8B & 88.66 & 87.02 \\
      GigaWorld-Policy & 5B & 86.36 & 85.04 \\
      Being-H0.7 & 3B & 90.20 & \textbf{89.60} \\
      \midrule
      Ours & 0.5B & \textbf{90.48} & 89.04 \\
      \bottomrule
    \end{tabular}
    \caption{Average Success rates (\%) over 50 RoboTwin~2.0 tasks.}
    \label{tab:robotwin}
\end{minipage}
\hfill
\begin{minipage}[t]{0.55\textwidth}
    \centering
    \small
    \setlength{\tabcolsep}{1mm}
    \begin{tabular}{llccccc}
      \toprule
      Method & Params & Spatial & Object & Goal & Long & AVG \\
      \midrule
      OpenVLA & 7B & 84.7 & 88.4 & 79.2 & 53.7 & 76.5 \\
      OpenVLA-OFT & 7B & 97.6 & 98.4 & \textbf{97.9} & 94.5 & \textbf{97.1} \\
      StarVLA & 4B & 97.8 & 98.6 & 96.2 & 93.8 & 96.6 \\
      $\pi_0$ & 3B & 96.8 & \textbf{98.8} & 95.8 & 85.2 & 94.1 \\
      GR00T-N1.6 & 3.3B & 97.7 & 98.5 & 97.5 & 94.4 & 97.0 \\
      JEPA-VLA & -- & 97.2 & 98.0 & 95.6 & \textbf{94.8} & 96.4 \\
      \midrule
      SmolVLA & 2B & 93.0 & 94.0 & 91.0 & 77.0 & 88.8 \\
      EVO-1 & 0.8B & 92.7 & 97.7 & 96.3 & 92.3 & 94.8 \\
      \midrule
      Ours & 0.5B & \textbf{98.0} & \textbf{98.8} & 97.2 & 94.2 & \textbf{97.1} \\
      \bottomrule
    \end{tabular}
    \caption{Success rates (\%) on the four LIBERO suites (Spatial, Object, Goal, and Long).}
    \label{tab:libero}
\end{minipage}
\end{table}

\paragraph{Results on RoboTwin 2.0.}
Table~\ref{tab:robotwin} compares LiLa-WAM with recent state-of-the-art
models~\cite{pi0,xvla,abotm0,pi05,starvla,gigaworld_policy,motus,beingh07}
on the 50 tasks of RoboTwin~2.0. Using a frozen DINOv3 encoder, LiLa-WAM is
trained on a single GPU and remains competitive with larger manipulation
policies. It surpasses the 8B Motus and 5B GigaWorld-Policy by 1.8 and 4.1
points, respectively, while using 16$\times$ and 10$\times$ fewer
parameters. In the clean setting, it slightly exceeds Being-H0.7 by 0.3
points.

\paragraph{Results on LIBERO.}
Table~\ref{tab:libero} compares LiLa-WAM with recent state-of-the-art
models~\cite{openvla,openvla_oft,starvla,pi0,groot_n1,jepa_vla,evo1,smolvla}
on LIBERO~\cite{libero}. At 0.5B parameters, LiLa-WAM attains a 97.1\%
average success rate, matching the baseline OpenVLA-OFT with roughly
14$\times$ fewer parameters and outperforming $\pi_0$ by 3.0 points. Among
lightweight models, it surpasses EVO-1 and SmolVLA by 2.3 and 8.3 points. These results suggest that foresight-aware manipulation can be realized with a lightweight, end-to-end trainable architecture under a single-GPU budget.

\subsection{Ablation Study}

\paragraph{Foresight supervision, task conditioning, and backbone.}
Table~\ref{tab:main_ablation} presents three comparisons. First, removing
the foresight objective $\mathcal{L}_{\mathrm{ff}}$ collapses the model to
a reactive VLA and drops the success rate from 70.0\% to 54.4\%, indicating that foresight supervision is an important contributor to performance. Second,
replacing the VTT with CLIP-encoded language instructions degrades
performance by 8.6 points, showing that the VTT offers
a viable lightweight form of task specification. Third, a Qwen3VL-2B backbone reaches 61.0\% despite being over 4$\times$
larger, showing that a larger pretrained backbone alone does not guarantee
better control performance under the same training budget.

\begin{table}[h]
\centering
\begin{tabular}{lcccc}
  \toprule
  Backbone & Foresight & Task Cond. & Params & AVG \\
  \midrule
  DINOv3 & $\times$ & VTT & 0.5B & 54.4 \\
  DINOv3 & $\checkmark$ & CLIP-lang & 0.5B & 61.4 \\
  Qwen3VL & $\checkmark$ & Language & 2.2B & 61.0 \\
  \midrule
  DINOv3 & $\checkmark$ & VTT & 0.5B & \textbf{70.0} \\
  \bottomrule
\end{tabular}
\caption{{Ablation on foresight supervision, task conditioning, and
visual backbone.} Average success rate over the 10
RoboTwin~2.0 tasks. The last row is our full model. Foresight supervision
and the VTT each contribute substantial gains, and the lightweight DINOv3 backbone outperforms a larger VLM backbone.}
\label{tab:main_ablation}
\end{table}

\paragraph{Number of adapter queries.}
Table~\ref{tab:query_num} reports the effect of the adapter query count
$N_q$, with $N_q = 64$ performing best (70.0\%). Smaller budgets (8 or 32)
likely lose spatial detail needed for fine-grained manipulation, whereas 96
tokens slightly degrades performance while increasing memory, suggesting
diminishing returns beyond a sufficient latent size. Training memory grows
steadily from 14.7\,GB to 21.3\,GB as $N_q$ increases. We thus adopt
$N_q = 64$, combining the best performance with a comfortable margin under
the 24\,GB budget.

\begin{table}[h]
\centering
\begin{tabular}{lcccc}
  \toprule
  $N_q$ (Adapter) & 8 & 32 & 64 & 96 \\
  \midrule
  AVG & 66.2 & 65.2 & \textbf{70.0} & 67.6 \\
  Training Memory (GB) & 14.7 & 16.0 & 18.0 & 21.3 \\
  \bottomrule
\end{tabular}
\caption{Effect of adapter query count $N_q$ on success rate and training memory on 10 RoboTwin~2.0 tasks.}
\label{tab:query_num}
\end{table}

\paragraph{Choice of DINOv3 feature layers.}
Table~\ref{tab:feat_layer} examines which encoder layers provide the most
control-relevant features. Performance improves steadily as features are
drawn from earlier blocks, rising from 48.0\% with the final block to
67.2\% with the 17th. A likely explanation is that later blocks emphasize
global semantics at the expense of spatial detail, whereas mid-level blocks
better preserve the object locations and local geometry that fine-grained
manipulation requires. Combining the 17th and 21st blocks further improves
the success rate to 70.0\%, indicating that the two levels offer
complementary cues.

\begin{table}[h]
\centering
\begin{tabular}{lccccc}
  \toprule
  Feature Layers & 24 & 23 & 21 & 17 & 17 \& 21 \\
  \midrule
  Avg & 48.0 & 53.2 & 57.4 & 67.2 & \textbf{70.0} \\
  \bottomrule
\end{tabular}
\caption{Effect of DINOv3 feature layers. Average success rate over
the 10 RoboTwin~2.0 tasks when aggregating different feature layers of the encoder.}
\label{tab:feat_layer}
\end{table}

\begin{figure}[h]
\centering
\includegraphics[width=0.65\columnwidth, trim=5 180 145 5,clip]{ 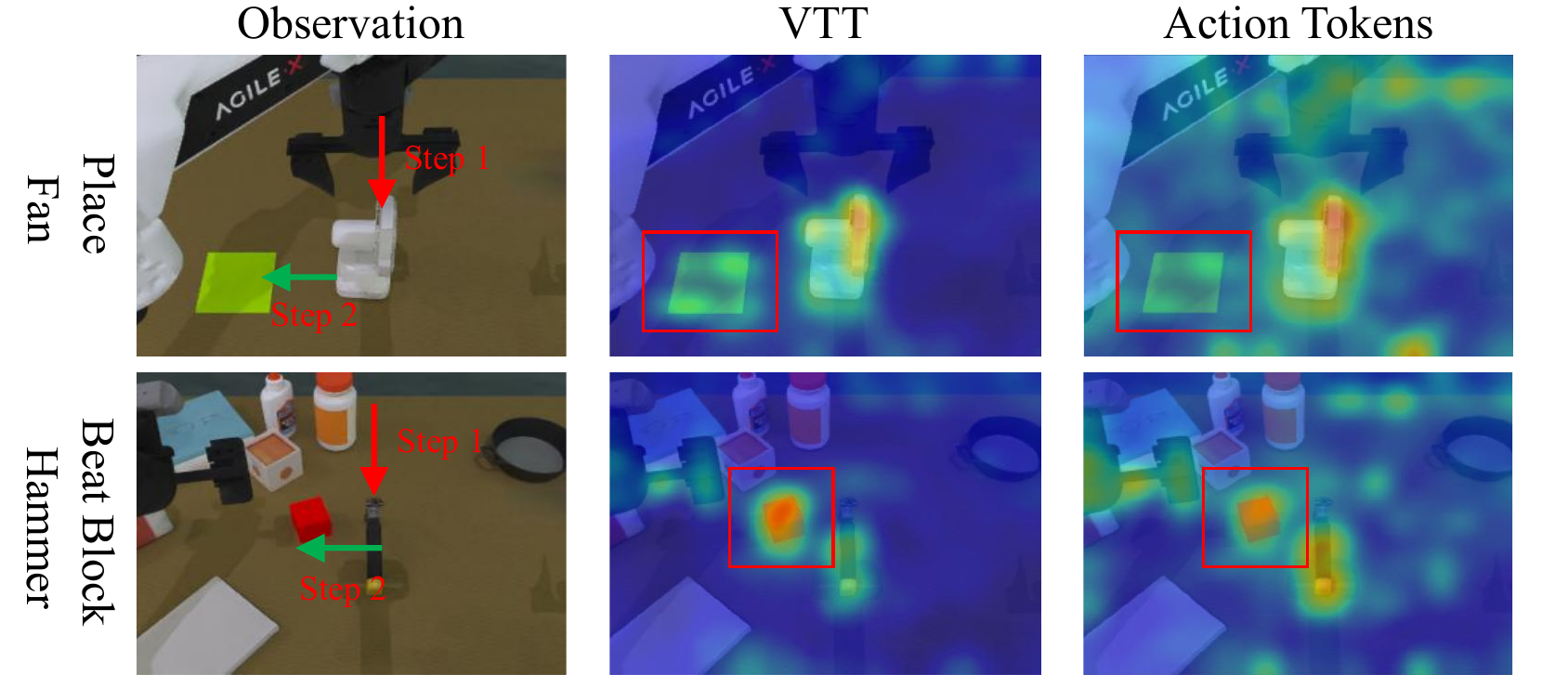}
\caption{Attention maps of the VTT and action tokens.}
\label{fig:AttnVis}
\end{figure}

\subsection{What Do the VTT and Action Tokens Attend To?}

\paragraph{Attention analysis.}
Figure~\ref{fig:AttnVis} compares the attention distributions of the VTT and
the action tokens on the input image. Interestingly, the two token types
exhibit a division of labor: the VTT focuses more on the objects
that the task acts upon (e.g., the fan and its target pad, the block and the
hammer), while the action tokens attend more to the arm and gripper,
with responses spreading over the manipulation region. This pattern matches
the intuition that task specification should capture what to
manipulate, whereas action generation track how the manipulator
moves, suggesting that the VTT provides task-relevant cues.

\begin{figure*}[t]
\centering
\includegraphics[width=0.9\textwidth, trim=20 84 20 5,clip]{ 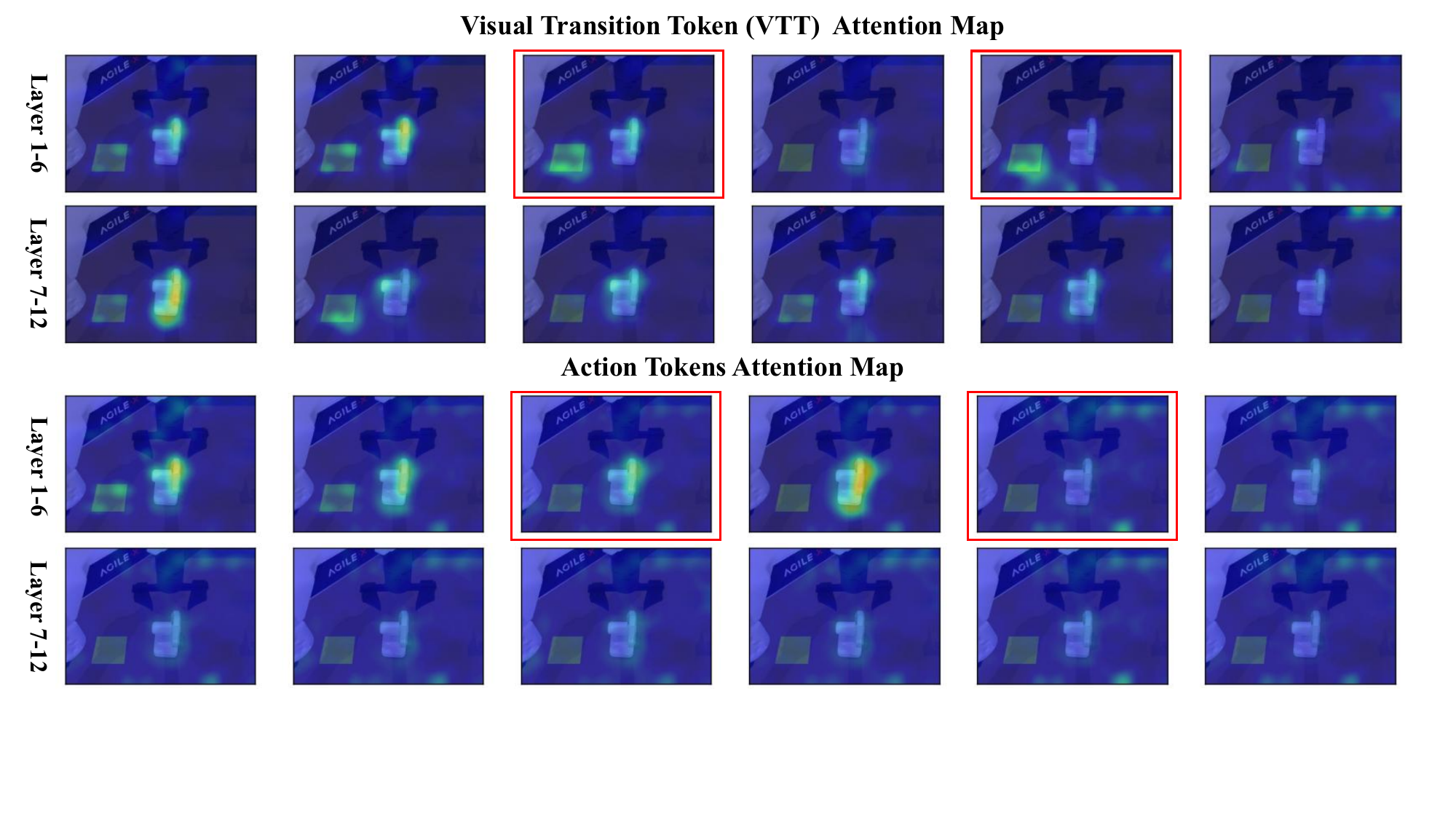} 
\caption{Per-layer attention maps of the VTT and action tokens. Compared
with the action tokens, the VTT attends more to the placement
area, whereas the action tokens attend more to the object being manipulated
(red boxes).}
\label{fig:AttnVisLayer1}
\end{figure*}

\paragraph{Per-layer attention analysis.}
Figure~\ref{fig:AttnVisLayer1} visualizes, for each of the $L{=}12$ DiT blocks of
the action expert, how the VTT and the action tokens distribute their
attention over the image. The visualization is conducted on the \emph{place
fan} task from RoboTwin~2.0, which requires grasping a small fan and placing
it onto a marked pad. We extract the attention weights from each token type
to the visual tokens in every block; since inference involves multiple
denoising steps, the action-token maps are averaged over all denoising steps.
Two observations emerge. First, the VTT activates on both the fan
to be grasped and the pad marking its destination. 
Second, the action tokens concentrate more
narrowly on the fan currently being manipulated. 
Together, the two token types exhibit a complementary division of labor: the VTT provides task-level context, while the action tokens focus on the object at hand, indicating that the VTT provides task-level guidance.

\subsection{Action Conditioning in the Action Expert}

\begin{figure}[h]
\centering
\includegraphics[width=0.65\columnwidth, trim=5 25 5 30,clip]{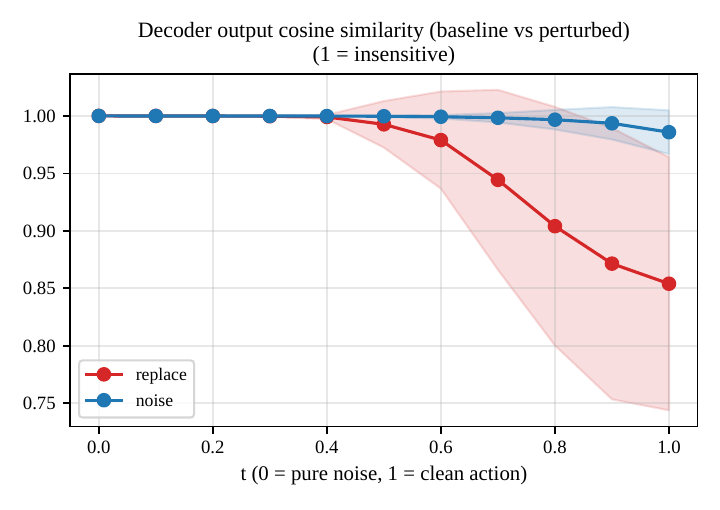}
\caption{Action perturbation probe of future feature prediction.
We perturb the action tokens at different denoising timesteps either by adding Gaussian noise or by replacing the entire action chunk with an action sequence sampled from another trajectory. The decoder-reconstructed future-feature tokens are used as the readout, and we report the cosine similarity between predictions obtained with the original and perturbed actions.}
\label{fig:action_perturbation}
\end{figure}

\paragraph{Action perturbation probe}
To examine whether the predicted future-state tokens encode action-conditioned dynamics, we intervene on the action tokens along the flow-matching trajectory. At each denoising timestep $s$, we either inject additional Gaussian noise into the action tokens or replace the whole action chunk with an action sequence sampled from another trajectory. Since the future-state tokens are compressed by the adapter, we use the decoder-reconstructed tokens as the readout and measure the cosine similarity between predictions obtained with the original and perturbed actions. A similarity of 1 indicates that the prediction is insensitive to the intervention. As shown in Figure~\ref{fig:action_perturbation}, both perturbations have little effect during the early denoising stage, where the action tokens are still dominated by noise. In contrast, replacing the action chunk leads to a clear similarity drop after $s \approx 0.5$, reaching approximately $0.85$ at $s=1$, whereas adding noise only causes a marginal decrease. This suggests that the predicted future features are not determined solely by the visual context, but encode action-conditioned future dynamics. The stronger sensitivity at late denoising stages is expected: as denoising progresses, the action tokens become semantically specific and the model gradually commits to a particular action trajectory, so replacing them introduces an inconsistent future. We further observe that the attention from visual-token positions to action tokens remains low across different denoising stages. This indicates that the late-stage sensitivity is not caused by a small number of high-attention connections, but rather by distributed information mixing through self-attention.

\paragraph{Attention mask analysis.}
Table~\ref{tab:attention_mask} reports the effect of suppressing the attention from visual-context tokens to action tokens on 10 RoboTwin~2.0 tasks. Under the default bidirectional self-attention, visual-context tokens can use action tokens as keys and values, allowing future feature prediction to obtain action-related information. When this attention pathway is suppressed, the average success rate decreases from 70.0\% to 64.8\%. This suggests that the action-to-visual information flow is important for dynamics-aware prediction, providing additional evidence that the predicted future features are action-conditioned. Notably, this degradation occurs even though the corresponding attention weights remain relatively low across denoising steps, indicating that useful action information may be transmitted through sparse or distributed attention patterns rather than a small number of dominant connections.

\begin{table}[h]
\centering
\small
\begin{tabular}{lcc}
  \toprule
  Attention Setting & Masked Query--KV  & Avg. Success (\%) \\
  \midrule
  Bidirectional attention & -- & \textbf{70.0} \\
  Visual-to-action attention mask & Q: $\mathbf{Z}_v$, K/V: $\mathbf{Z}_a$ & 64.8 \\
  \bottomrule
\end{tabular}
\caption{Effect of suppressing the attention from visual-context tokens to action tokens on 10 RoboTwin~2.0 tasks. $\mathbf{Z}_v$ denotes the compressed visual-context tokens, and $\mathbf{Z}_a$ denotes the action tokens. }
\label{tab:attention_mask}
\end{table}

\begin{figure}[h]
\centering
\includegraphics[width=0.65\columnwidth, trim=5 6 5 5,clip]{ 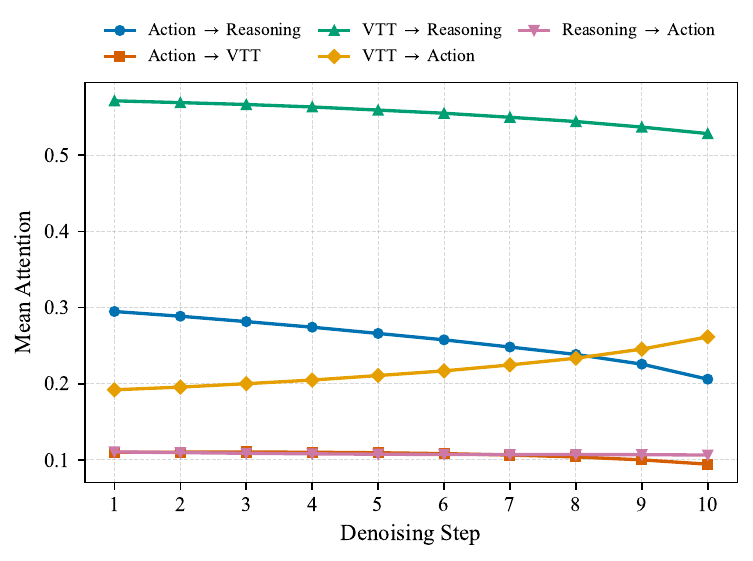}
\caption{Attention dynamics across denoising steps on the \emph{press
stapler} task from RoboTwin~2.0. Mean attention between token groups at
each of the 10 flow-matching denoising steps, averaged over all DiT blocks.}
\label{fig:attn_press_stapler}
\end{figure}

\paragraph{Attention dynamics across denoising steps.}
The Foresight-Aware Action Expert applies bidirectional self-attention over
the full token sequence without any attention mask. All token groups can
thus freely exchange information. Figure~\ref{fig:attn_press_stapler}
tracks the mean attention between token groups across the denoising steps, averaged over all DiT blocks.

Three trends stand out. First, the VTT consistently assigns most of its
attention to the reasoning tokens throughout denoising. 
This suggests that the task representation maintains interaction with the visual context throughout denoising. 
Second, the action tokens attend
strongly to the reasoning tokens at early steps. This reliance gradually
decreases as denoising proceeds, which may reflect a transition from context gathering to action refinement.
Third, attention from the reasoning tokens to the action tokens remains
low throughout denoising.

\subsection{Real-Robot Experiments}
\label{sec:exp_real}

\begin{figure}[h]
\centering
\includegraphics[width=0.8\columnwidth, trim=5 254 5 5,clip]{ 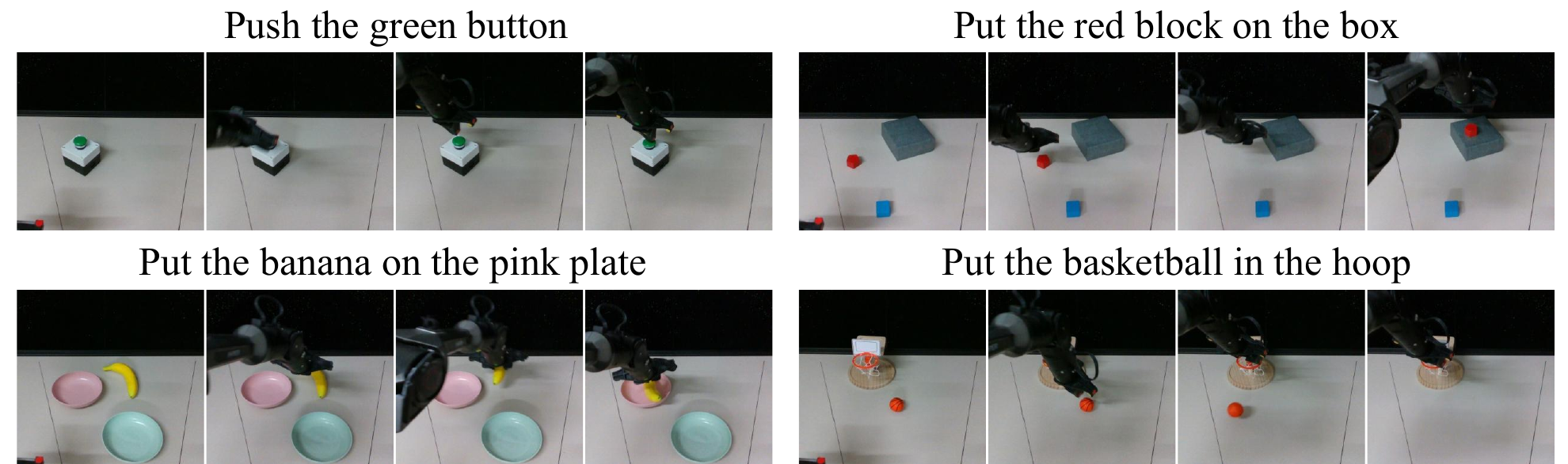}
\caption{Execution process of the four real-robot tasks.}
\label{fig:RealRobotTaskVis}
\end{figure}

\paragraph{Setup.}
As shown in Figure~\ref{fig:RealRobotTaskVis}, we conduct real-robot
experiments on an Agilex Piper 6-DoF robotic arm, using an Intel RealSense
D435 camera as the third-person view. The policy receives RGB observations
resized to $256\times256$ and outputs absolute joint-angle commands at
30\,Hz. The four tasks in Table~\ref{tab:real_robot} are evaluated with
randomized object placements to test robustness to variations in the initial
scene configuration. For each task, we collect 60 demonstrations and train
both variants on the same data. Each method is evaluated over 50 rollouts
per task, and we report the success rate. We compare LiLa-WAM with its
variant without the foresight loss $\mathcal{L}_{\mathrm{ff}}$, which
isolates the contribution of future-state supervision on the physical robot.
More setup details are provided in the supplementary material.

\paragraph{Results.}
Table~\ref{tab:real_robot} reports the real-robot results. LiLa-WAM
achieves higher success rates than the ablated variant on all four tasks,
improving the average success rate from 74.0\% to 82.0\%. The gains range
from 4 to 14 points, with the largest improvement observed on
\emph{push the green button}. \emph{Put the basketball in the hoop} remains
the most challenging task for both variants. These results indicate that
foresight supervision also benefits real-world robotic control, although
performance on more difficult tasks still leaves room for improvement.

\begin{table}[h]
\centering
\setlength{\tabcolsep}{2mm}
\begin{tabular}{lccccc}
  \toprule
  Method & Button & Block & Banana & Basketball & Avg \\
  \midrule
  w/o $\mathcal{L}_{\mathrm{ff}}$ & 72 & 84 & 92 & 48 & 74.0 \\
  LiLa-WAM & \textbf{86} & \textbf{90} & \textbf{96} & \textbf{56} & \textbf{82.0} \\
  \bottomrule
\end{tabular}
\caption{Real-robot success rates (\%). The four tasks are \emph{push the green button} (Button), \emph{put the red block on the box} (Block), \emph{put the banana on the pink plate} (Banana), and \emph{put the basketball in the hoop} (Basketball).}
\label{tab:real_robot}
\end{table}

\section{Conclusion}
\label{sec:conclusion}

In this paper, we propose LiLa-WAM, a lightweight world-action model that
reasons about the future in a compact latent space jointly shaped by
future-state prediction and action generation, and can be trained end-to-end
on a single GPU. We further introduce the Visual Transition Token, a
language-free task representation that encodes each task as a direction in
visual feature space. Experiments on RoboTwin~2.0, LIBERO, and
real-robot tasks, together with ablation and visualization analyses,
support the effectiveness of the proposed model.

\bibliographystyle{unsrtnat}   
\bibliography{references}

\newpage
\appendix

\begin{table*}[t]
\centering

\setlength{\tabcolsep}{3pt}
\renewcommand{\arraystretch}{0.95}
\begin{tabular}{lcc ccc ccc cc}
  \toprule
  & \multicolumn{2}{c}{$\pi_{0.5}$}
  & \multicolumn{2}{c}{Motus}
  & \multicolumn{2}{c}{X-VLA}
  & \multicolumn{2}{c}{GigaWorld-Policy}
  & \multicolumn{2}{c}{LiLa-WAM} \\
  \cmidrule(lr){2-3}
  \cmidrule(lr){4-5}
  \cmidrule(lr){6-7}
  \cmidrule(lr){8-9}
  \cmidrule(lr){10-11}
  Task & Clean & Rand. & Clean & Rand. & Clean & Rand. & Clean & Rand. & Clean & Rand. \\
  \midrule
  Adjust Bottle & 100 & 99 & 89 & 93 & 100 & 99 & 100 & 100 & 100 & 98 \\
  Beat Block Hammer & 96 & 93 & 95 & 88 & 92 & 88 & 86 & 86 & 96 & 90 \\
  Blocks Ranking RGB & 92 & 85 & 99 & 97 & 83 & 83 & 92 & 96 & 100 & 100 \\
  Blocks Ranking Size & 49 & 26 & 75 & 63 & 67 & 74 & 44 & 48 & 92 & 88 \\
  Click Alarmclock & 98 & 89 & 100 & 100 & 99 & 99 & 100 & 100 & 94 & 92 \\
  Click Bell & 99 & 66 & 100 & 100 & 100 & 100 & 100 & 100 & 90 & 90 \\
  Dump Bin Bigbin & 92 & 97 & 95 & 91 & 79 & 77 & 92 & 100 & 96 & 94 \\
  Grab Roller & 100 & 100 & 100 & 100 & 100 & 100 & 100 & 100 & 100 & 100 \\
  Handover Block & 66 & 57 & 86 & 73 & 73 & 37 & 80 & 80 & 96 & 90 \\
  Handover Mic & 98 & 97 & 78 & 63 & 0 & 0 & 72 & 72 & 100 & 98 \\
  Hanging Mug & 18 & 17 & 38 & 38 & 23 & 27 & 16 & 12 & 56 & 44 \\
  Lift Pot & 96 & 85 & 96 & 99 & 99 & 100 & 98 & 98 & 100 & 96 \\
  Move Can Pot & 51 & 55 & 34 & 74 & 89 & 86 & 76 & 78 & 98 & 90 \\
  Move Pillbottle Pad & 84 & 61 & 93 & 96 & 73 & 71 & 90 & 90 & 98 & 94 \\
  Move Playingcard Away & 96 & 84 & 100 & 96 & 93 & 98 & 78 & 72 & 100 & 100 \\
  Move Stapler Pad & 56 & 42 & 83 & 85 & 78 & 73 & 92 & 82 & 66 & 76 \\
  Open Laptop & 90 & 96 & 95 & 91 & 93 & 100 & 96 & 98 & 94 & 100 \\
  Open Microwave & 34 & 77 & 95 & 91 & 79 & 71 & 74 & 66 & 66 & 58 \\
  Pick Diverse Bottles & 81 & 71 & 90 & 91 & 58 & 36 & 82 & 70 & 84 & 88 \\
  Pick Dual Bottles & 93 & 63 & 96 & 90 & 47 & 36 & 86 & 86 & 100 & 100 \\
  Place A2B Left & 87 & 82 & 88 & 79 & 48 & 49 & 94 & 88 & 68 & 78 \\
  Place A2B Right & 87 & 84 & 91 & 87 & 36 & 36 & 90 & 92 & 76 & 76 \\
  Place Bread Basket & 77 & 64 & 91 & 94 & 81 & 71 & 82 & 82 & 94 & 98 \\
  Place Bread Skillet & 85 & 66 & 86 & 83 & 77 & 67 & 94 & 90 & 92 & 88 \\
  Place Burger Fries & 94 & 87 & 98 & 98 & 94 & 94 & 98 & 96 & 96 & 96 \\
  Place Can Basket & 62 & 62 & 81 & 76 & 49 & 52 & 78 & 74 & 80 & 72 \\
  Place Cans Plasticbox & 94 & 84 & 98 & 94 & 97 & 98 & 100 & 100 & 98 & 100 \\
  Place Container Plate & 99 & 95 & 98 & 99 & 97 & 95 & 98 & 96 & 96 & 98 \\
  Place Dual Shoes & 75 & 75 & 93 & 87 & 79 & 88 & 96 & 84 & 60 & 54 \\
  Place Empty Cup & 100 & 99 & 99 & 98 & 100 & 98 & 90 & 90 & 100 & 100 \\
  Place Fan & 87 & 85 & 91 & 87 & 80 & 75 & 92 & 94 & 96 & 90 \\
  Place Mouse Pad & 60 & 39 & 66 & 68 & 70 & 70 & 88 & 90 & 88 & 76 \\
  Place Object Basket & 80 & 76 & 81 & 87 & 44 & 39 & 90 & 92 & 88 & 90 \\
  Place Object Scale & 86 & 80 & 88 & 85 & 52 & 74 & 88 & 80 & 96 & 88 \\
  Place Object Stand & 91 & 85 & 98 & 97 & 86 & 88 & 100 & 98 & 92 & 94 \\
  Place Phone Stand & 81 & 81 & 87 & 86 & 88 & 87 & 82 & 72 & 92 & 94 \\
  Place Shoe & 92 & 93 & 99 & 97 & 96 & 95 & 98 & 96 & 100 & 98 \\
  Press Stapler & 87 & 83 & 93 & 98 & 92 & 98 & 96 & 96 & 100 & 98 \\
  Put Bottles Dustbin & 84 & 79 & 81 & 79 & 74 & 77 & 72 & 70 & 92 & 94 \\
  Put Object Cabinet & 80 & 79 & 88 & 71 & 46 & 48 & 74 & 74 & 92 & 92 \\
  Rotate QRcode & 89 & 87 & 89 & 73 & 34 & 33 & 90 & 84 & 88 & 84 \\
  Scan Object & 72 & 65 & 67 & 66 & 14 & 36 & 60 & 64 & 94 & 90 \\
  Shake Bottle & 99 & 97 & 100 & 97 & 99 & 100 & 100 & 100 & 100 & 100 \\
  Shake Bottle Horizontally & 99 & 99 & 100 & 98 & 100 & 100 & 100 & 98 & 100 & 100 \\
  Stack Blocks Three & 91 & 76 & 91 & 95 & 6 & 10 & 70 & 78 & 86 & 78 \\
  Stack Blocks Two & 97 & 100 & 100 & 98 & 92 & 87 & 100 & 94 & 100 & 98 \\
  Stack Bowls Three & 77 & 71 & 79 & 87 & 76 & 86 & 70 & 72 & 88 & 82 \\
  Stack Bowls Two & 95 & 96 & 98 & 98 & 96 & 93 & 96 & 92 & 98 & 100 \\
  Stamp Seal & 79 & 55 & 93 & 92 & 76 & 82 & 96 & 98 & 72 & 78 \\
  Turn Switch & 62 & 54 & 84 & 78 & 40 & 61 & 82 & 84 & 76 & 82 \\
  \midrule
  Average & 82.74 & 76.76 & 88.66 & 87.02 & 72.88 & 72.84 & 86.36 & 85.04 & \textbf{90.48} & \textbf{89.04} \\
  \bottomrule
\end{tabular}
\caption{Per-task success rates (\%) on 50 RoboTwin~2.0 tasks under clean and randomized settings.}
\label{tab:robotwin_per_task}
\end{table*}

\section{Real-Robot Experiment Details}

\paragraph{Hardware platform.}
Real-robot experiments are conducted on an AgileX PiPer robotic arm, a
6-DoF manipulator equipped with a parallel gripper. The action space
consists of absolute joint positions, and the gripper is controlled with
continuous values. A RealSense D435 camera provides RGB observations from a
third-person viewpoint, mounted at a head-like position overlooking the
workspace. The policy predicts action chunks of size 32, and the robot
operates at a control frequency of 30\,Hz.

\paragraph{Asynchronous inference.}
To prevent execution pauses caused by inference latency, we adopt an
asynchronous inference scheme with multi-threading. A dedicated thread
maintains a queue of actions to be executed. When the number of remaining
actions in the queue drops to 8, model inference is triggered in a separate
thread; upon completion, the queue is refreshed with the newly predicted
action chunk. In this way, the robot always has pending actions to execute
while the model is inferring, which eliminates pauses and ensures the
continuity of motion.
All real-robot experiments, including the baseline without future-state
supervision, use identical data preprocessing and postprocessing pipelines.
This controlled setting helps isolate the effect of future-state supervision from differences in data handling.

\begin{figure}[h]
\centering
\includegraphics[width=0.7\columnwidth, trim=5 245 430 5,clip]{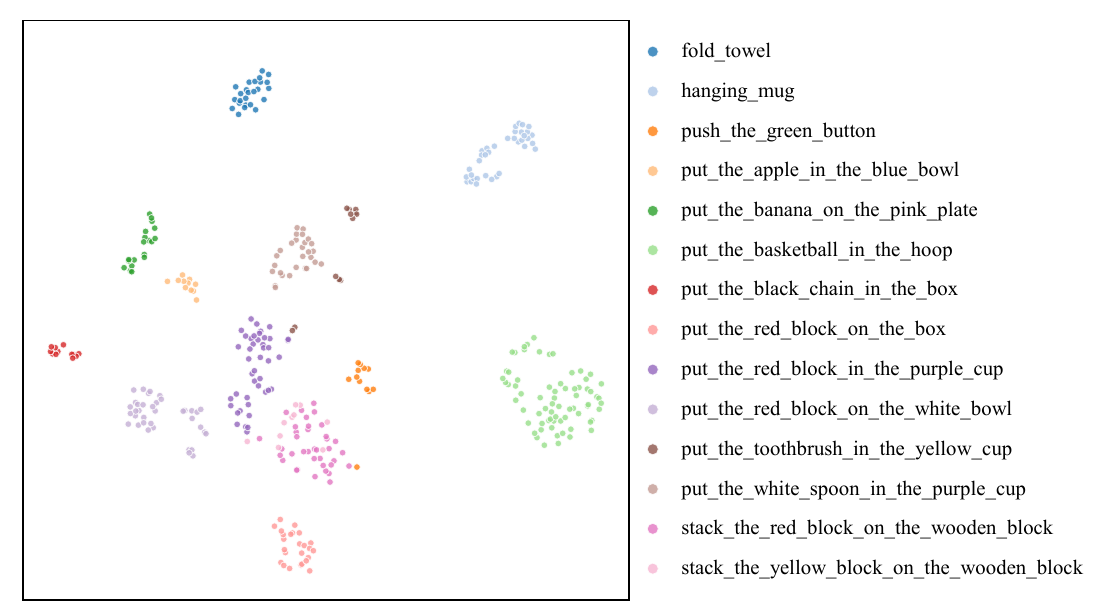}
\caption{t-SNE visualization of per-episode visual transition embeddings on the real-robot demonstration set, where each point corresponds to one episode.}
\label{fig:TsneRealRobot}
\end{figure}

\begin{figure}[h]
\centering
\includegraphics[width=0.7\columnwidth, trim=5 305 490 5,clip]{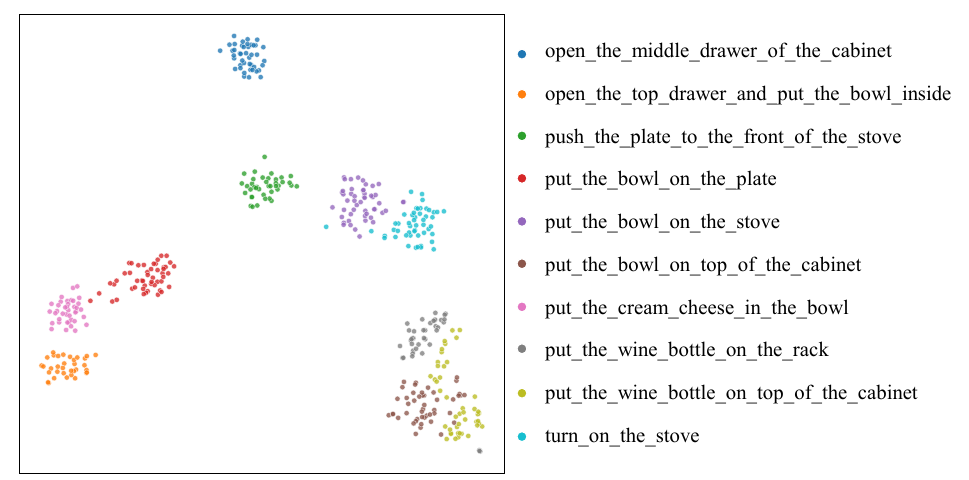}
\caption{t-SNE visualization of per-episode visual transition embeddings on LIBERO-goal.
Each point represents the visual transition embedding of a single episode. Semantically related tasks that share objects or scenes (e.g., \emph{put the wine bottle on top of the cabinet} and \emph{put the wine bottle on the rack}) tend to have closer clusters, suggesting that the visual transition embeddings capture both task-relevant transition information and task similarity.}
\label{fig:TsneLiberoGoal}
\end{figure}

\end{document}